%% file: main.tex
\documentclass[runningheads]{llncs}

\usepackage{eccv}

\usepackage{eccvabbrv}

\usepackage{graphicx}
\usepackage{booktabs}
\usepackage{multicol}
\usepackage{multirow}
\usepackage{bbding}
\usepackage{makecell}
\usepackage{tablefootnote}
\usepackage[accsupp]{axessibility}  
\def\Name{TFVTG}

\usepackage{hyperref}

\usepackage{orcidlink}

\begin{document}

\title{Training-free Video Temporal Grounding using Large-scale Pre-trained Models} 

\titlerunning{Training-free Video Temporal Grounding}

\author{Minghang Zheng\inst{1}\orcidlink{0000-0003-1612-975X} \and
Xinhao Cai\inst{1}\orcidlink{0009-0009-5459-3458} \and
Qingchao Chen\inst{2}\orcidlink{0000-0002-1216-5609} \and
Yuxin Peng\inst{1}\orcidlink{0000-0001-7658-3845} \and
Yang Liu\orcidlink{0000-0002-4259-3882} \inst{1,3}\thanks{Corresponding author}}

\authorrunning{M.~Zheng et al.}

\institute{Wangxuan Institute of Computer Technology, Peking University \and
National Institute of Health Data Science, Peking University \and
State Key Laboratory of General Artificial Intelligence, Peking University\\
\email{\{minghang,qingchao.chen,pengyuxin,yangliu\}@pku.edu.cn}
\email{xinhao.cai@stu.pku.edu.cn}}

\maketitle

\begin{abstract}
  Video temporal grounding aims to identify video segments within untrimmed videos that are most relevant to a given natural language query. Existing video temporal localization models rely on specific datasets for training, with high data collection costs, but exhibit poor generalization capability under the across-dataset and out-of-distribution (OOD) settings. In this paper, we propose a \textbf{T}raining-\textbf{F}ree \textbf{V}ideo \textbf{T}emporal \textbf{G}rounding (\Name{}) approach that leverages the ability of pre-trained large models. A naive baseline is to enumerate proposals in the video and use the pre-trained visual language models (VLMs) to select the best proposal according to the vision-language alignment. However, most existing VLMs are trained on image-text pairs or trimmed video clip-text pairs, making it struggle to (1) grasp the relationship and distinguish the temporal boundaries of multiple events within the same video; (2) comprehend and be sensitive to the dynamic transition of events (the transition from one event to another) in the video. To address these issues, firstly, we propose leveraging large language models (LLMs) to analyze multiple sub-events contained in the query text and analyze the temporal order and relationships between these events. Secondly, we split a sub-event into dynamic transition and static status parts and propose the dynamic and static scoring functions using VLMs to better evaluate the relevance between the event and the description. Finally, for each sub-event description provided by LLMs, we use VLMs to locate the top-k proposals that are most relevant to the description and leverage the order and relationships between sub-events provided by LLMs to filter and integrate these proposals.  Our method achieves the best performance on zero-shot video temporal grounding on Charades-STA and ActivityNet Captions datasets without any training and demonstrates better generalization capabilities in cross-dataset and OOD settings. Code is available at \url{https://github.com/minghangz/TFVTG}.
  
\end{abstract}

\input{secs/1_intro}
\input{secs/2_related}
\input{secs/3_method}
\input{secs/4_experiments}
\input{secs/5_conclusion}

%
%
\bibliographystyle{splncs04}
\bibliography{main}
\end{document}

%% file: secs/1_intro.tex
\begin{figure}[th]
    \centering
    \includegraphics[width=\linewidth]{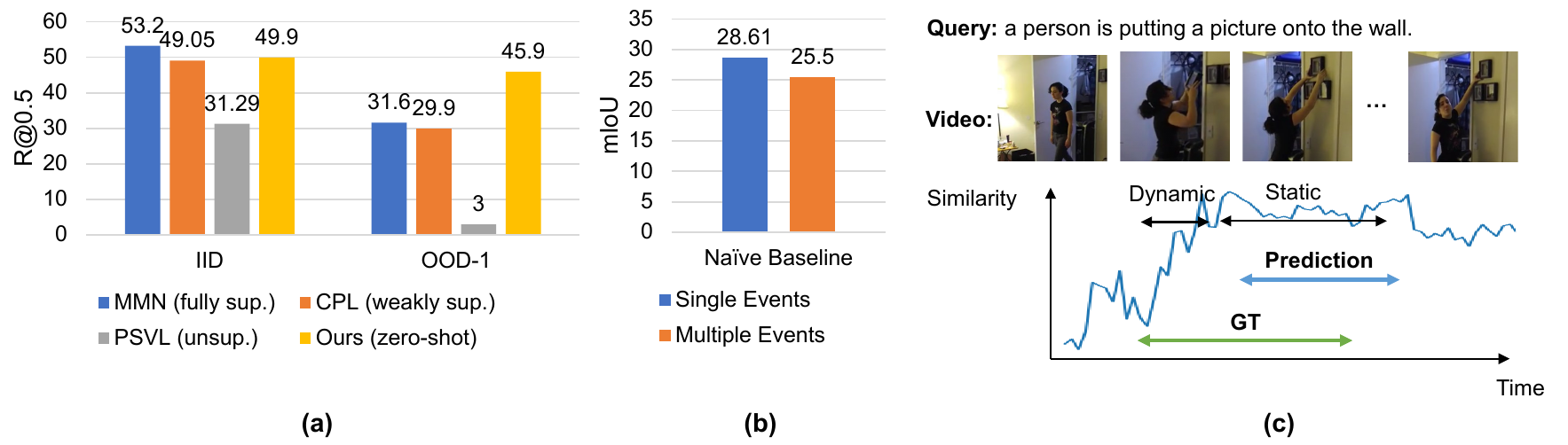}
    \caption{(a) Evaluation results of existing methods and our method under the IID and OOD setting on the Charades-STA dataset. (b) Evaluation results of the naive baseline on the ActivityNet Datasets when the query describes single or multiple events. (c) The query-frame similarity obtained from the BLIP-2 Q-Former. The naive baseline based on BLIP-2 tends to predict the static parts of the video and ignores the dynamic transitions.}
    \label{fig:teaser}
\end{figure}

\section{Introduction}

Video temporal grounding~\cite{gao2017tall} aims to localize the most semantically relevant segment in untrimmed videos according to free-form natural language queries. It has broad application potential in video surveillance~\cite{collins2000system}, video summarization~\cite{10.1145/641007.641116}, and other fields. 
Existing video temporal grounding methods~\cite{gao2017tall,Wang_2021_CVPR,Zhao_2021_CVPR,zhou2021embracing,huang2022video,zhang2020learning,zhang2021multi,TRM_2023_AAAI,luo2023vid,jang2023EaTR} mainly train models on manually annotated data to understand the alignment between video segments and natural language queries. These methods have achieved remarkable improvements recently on specific datasets, such as ActivityNet Captions~\cite{2017Dense} and Charades-STA~\cite{gao2017tall}. However, collecting high-quality video temporal grounding datasets is time-consuming and labor-intensive, which hinders the large-scale real-world application of these methods. In addition, they heavily rely on the distribution of the training dataset and the performance degrades significantly in out-of-distribution (OOD) or cross-dataset settings according to previous research~\cite{yuan2021closer, yang2021deconfounded, bao2022learning,li2022compositional}. As shown in Figure~\ref{fig:teaser} (a), we report the OOD performance~\footnote{We follow \cite{yang2021deconfounded} and plot the performance under the OOD-1 setting in the figure.}  on the Charades-STA dataset of recent fully supervised method MMN~\cite{MMN}, weakly supervised method CPL~\cite{CPL_2022_CVPR}, and unsupervised method PSVL~\cite{nam2021zero}. As we can see, their performance all has a significant drop. This is because these models are trained on small-scale video temporal grounding datasets that exhibit certain biases, leading to poor generalization of the models. Therefore, we aim to design a \textit{training-free video temporal grounding} approach that \textit{does not rely on specific video temporal grounding datasets}, so that it can be better generalized to real application scenarios.

Recently, large-scale pre-trained models~\cite{radford2021learning,li2022blip,zhang2023videollama,liu2024LLaVA,achiam2023gpt,team2023gemini,wang2023internvid,wang2022internvideo} have shown strong generalization ability in zero-shot retrieval~\cite{yelamarthi2018zero,luo2024zero}, VQA~\cite{guo2023images,mo2024bridging}, detection~\cite{yang2024active,lei2024exploring,lei2024exploring2} et al. This inspires us to transfer the powerful generalization ability of them to video temporal grounding tasks. A naive baseline is to enumerate proposals in the video and use the pre-trained visual language models (VLMs)~\cite{radford2021learning,li2022blip,wang2023internvid,wang2022internvideo} to assess the alignment between these proposals and the query and select the proposal with the highest score. However, this approach has the following drawbacks. \textit{Firstly, it can be challenging for VLMs to understand the temporal boundaries of multiple events in untrimmed video.}   
Most of the existing image-text or video-text pre-trained VLMs are trained to align single images (e.g. CLIP~\cite{radford2021learning}, BLIP~\cite{li2022blip}) or trimmed video clips (e.g. InterVideo~\cite{wang2022internvideo}) with texts. However, in the video temporal grounding, the model needs to understand multiple sequential events and their temporal relationships, such as `She sprays it with a spray bottle and continues brushing her hair'. The skill is seldom emphasized in the image or trimmed video pretraining, and as shown in Figure~\ref{fig:teaser}(b), the naive baseline has a poorer performance when the query describes multiple events.
\textit{Secondly, we find that directly selecting the most relevant proposal using VLMs often leads to overlooking the dynamic transitions at the beginning of an event.} As shown in Figure~\ref{fig:teaser}(c), we show the similarity between video frames and the query text using the BLIP-2 Q-Former~\cite{li2023blip}. It can be found that in the naive baseline's prediction, the beginning of the event where a person gradually picks up the picture and approaches the wall is ignored. 
We think this is because these VLMs are trained directly by visual-textual contrastive learning. In such a training paradigm, the model's primary objective is to associate the most discriminative visual cues with their corresponding text descriptions, rather than emphasizing the need to focus on dynamic transitions and ensure the completeness of localized event boundaries.

To address the above problems, we propose to combine the ability of large language models (LLMs)~\cite{achiam2023gpt,team2023gemini,liu2024LLaVA} to understand and reason about queries and the ability of visual language models (VLMs) to align vision and text in a training-free manner. 
Specifically, for the challenge of understanding videos and queries that contain multiple events, we propose to prompt the LLMs to analyze the multiple events that may be contained in the query and give a text description of each single event as sub-queries, the order in which each event occurs, and the relationship between events (sequential or simultaneous).
For the sub-query of each single event, we can use the VLMs to locate its possible occurrence in the video.
To better use the VLMs to locate the video clip corresponding to the single event query and solve the problem of ignoring the dynamic transitions in the video while enhancing localization completeness, we propose to consider both dynamic transition and static status after transition explicitly when measuring the similarity between the proposal and the text query. For example in Figure~\ref{fig:teaser}(c), for the query `a person put a picture on the wall', the dynamic transition part is where the person gradually raises the picture and approaches the wall, while the static status part is where the person has already hung the picture on the wall and is looking at the camera.
A good proposal should begin with a dynamic segment, exhibiting a notable increase in video-text similarity and followed by a static post-status segment characterized by a high average video-text similarity within the static segment, while displaying a low average similarity outside of it.
To evaluate each proposal, we compute a matching score by summing the dynamic score, which measures the rate of similarity change within its dynamic segment, and the static score, which evaluates the comparative similarity within and outside its static post-status segment.
Then, we select the top proposals with the highest matching scores as the localization results of the single event description. Finally, the localization results of each single event are combined with the LLM's judgment of the relationship and order of events to filter and integrate the final predictions.

Our contributions are summarized as follows. (1) We propose a training-free pipeline for video temporal grounding (\Name{}) using pre-trained large language models (LLMs) and vison-language models (VLMs). We use the LLMs to split the original query into sub-events and reason the temporal order and relationship between them, use VLMs to localize each sub-event, and filter and integrate the localization results based on the temporal order and relationships. (2) To help VLM better understand the dynamic transitions in the video, we divide the events into dynamic and static parts and model them separately. For the dynamic part, measures the rate of similarity change, and for the static part, we measure the comparative similarity within and outside. 
(3) Our method achieves the best performance on zero-shot video temporal grounding on both the Charades-STA~\cite{gao2017tall} and ActivityNet Captions~\cite{2017Dense} datasets and has a greater advantage in cross-dataset and OOD settings.

%% file: secs/2_related.tex
\section{Related Work}

\subsection{Video Temporal Localization}
Fully supervised video temporal grounding methods~\cite{zhang2020learning,gao2017tall,huang2022video,liu2022memory,jang2023EaTR,LGI,yang2021deconfounded,MMN,luo2023vid,croitoru2023moment} typically train models using manually annotated queries and start and end timestamps. For example, MMN~\cite{MMN} trains models to distinguish matched and unmatched video-sentence pairs collected from within videos and across videos; VTimeLLM~\cite{huang2023vtimellm} is the first to fine-tune large language models using video temporal grounding data. However, fully supervised methods are often influenced by annotation bias, leading to poor generalization. Weakly supervised learning~\cite{CPL_2022_CVPR,CNM_2022_AAAI,huang2023weakly} and unsupervised learning~\cite{nam2021zero,zheng2023SPL,ye2022unsupervised,li2024flowgananomaly} are often used to mitigate the high dependence on human annotation.
For instance, PSVL~\cite{nam2021zero} and SPL~\cite{zheng2023SPL} train models by generating pseudo-labels within videos. Nevertheless, even without using manually annotated data, biases present in the training videos can still affect the generalization of these methods. Therefore, in this work, we focus on training-free video temporal grounding, aiming for stronger generalization and applicability in real-world scenarios. Recently, Luo et al.~\cite{luo2024zero} and VTG-GPT~\cite{xu2024vtggpt} made the first attempt to training-free video temporal grounding. Luo et al. measure the similarity between video proposals and the text query using VLM while VTG-GPT measures the similarity between the video frame captions and the text query using LLM. They then make a prediction based on the similarity. However, they overlooked the order and relationship between the possible multiple events within the query, as well as the issue of VLM's insensitivity to dynamic transitions in the video due to their training scheme. In contrast, we propose to infer multiple sub-events contained in the query and their order and relationship through LLM, model dynamic changes in the video explicitly to assist VLM in better localizing each sub-event, and filter and integrate the localization results based on the order and relationship between events inferred by LLM. 

\subsection{Bias in Video Temporal Localization Datasets}

The mainstream video temporal grounding datasets~\cite{2017Dense, gao2017tall} suffer from certain biases, which affect the generalization ability of models trained on these datasets. Many studies~\cite{yuan2021closer,yang2021deconfounded,bao2022learning,li2022compositional} have investigated biases in video temporal grounding datasets. For example, \cite{yuan2021closer,yang2021deconfounded} study biases in the location of target segments. When there are significant changes in the distribution of locations, existing methods exhibit noticeable performance degradation. \cite{li2022compositional} explores biases in query texts and proposes the ActivityNet-CG and Charades-CG datasets to evaluate the generalization ability of existing methods in the novel combinations of phrases and novel words. \cite{bao2022learning} studies the cross-dataset generalization ability of existing models and finds that models trained on specific datasets perform poorly when testing on another dataset. Some works~\cite{li2022compositional,bao2022learning,yang2021deconfounded,yang2023deco} have focused on improving model generalization to specific biases. However, they are difficult to generalize to address various types of biases in video temporal grounding. Therefore, we attempt to study training-free video temporal grounding to overcome reliance on specific datasets and achieve better generalization across various scenarios.

\subsection{Large-scale Pretrained Models in Video Understanding}
In recent years, with the development of large-scale corpus~\cite{schuhmann2021laion,bain2021frozen}, model architecture~\cite{vaswani2017attention,tian2023transformer}, and computational resources, large language models (LLMs)~\cite{achiam2023gpt,team2023gemini,liu2024LLaVA} have achieved rapid development, demonstrating powerful capabilities in text generation, chat, problem-solving, reasoning, et al. On the other hand, visual language models (VLMs)~\cite{radford2021learning,li2022blip,wang2022internvideo,yang20243d} have also shown strong generalization abilities in tasks such as multimodal alignment and retrieval. Some existing methods attempt to combine VLMs with LLMs in video tasks. For example, ChatVideo~\cite{wang2023chatvideo} utilizes pre-trained models such as trajectory detection, video captioning, and speech recognition to convert videos into text, which serves as input to LLMs for video understanding. Video-LLaMA~\cite{zhang2023videollama}, and VTimeLLM~\cite{huang2023vtimellm} project video features into the token embedding space of LLMs through fine-tuning to enable LLMs to understand videos. However, these methods perform poorly on video temporal grounding tasks as shown in Table~\ref{tab:iid}. Even VTimeLLM, which specifically fine-tunes LLM using video temporal grounding data, still exhibits a significant performance gap compared to the traditional video temporal grounding method. We think that this may be because fine-tuning VLMs and LLMs on video temporal grounding datasets degrades their generalization, and encoding videos solely as input token sequences for LLMs makes it difficult for LLMs to accurately understand the time boundaries of various events. Therefore, we propose a training-free pipeline to combine the capabilities of LLMs and VLMs for video temporal grounding tasks. We leverage the strengths of both: using LLMs to reason the sub-events contained in queries, their occurrence order, and relationships, using VLMs to measure the vision-text similarity and localize each sub-event, and integrating the final predictions based on the inferred sub-event order and relationships from LLM.

%% file: secs/3_method.tex
\begin{figure}[th]
    \centering
    \includegraphics[width=\linewidth]{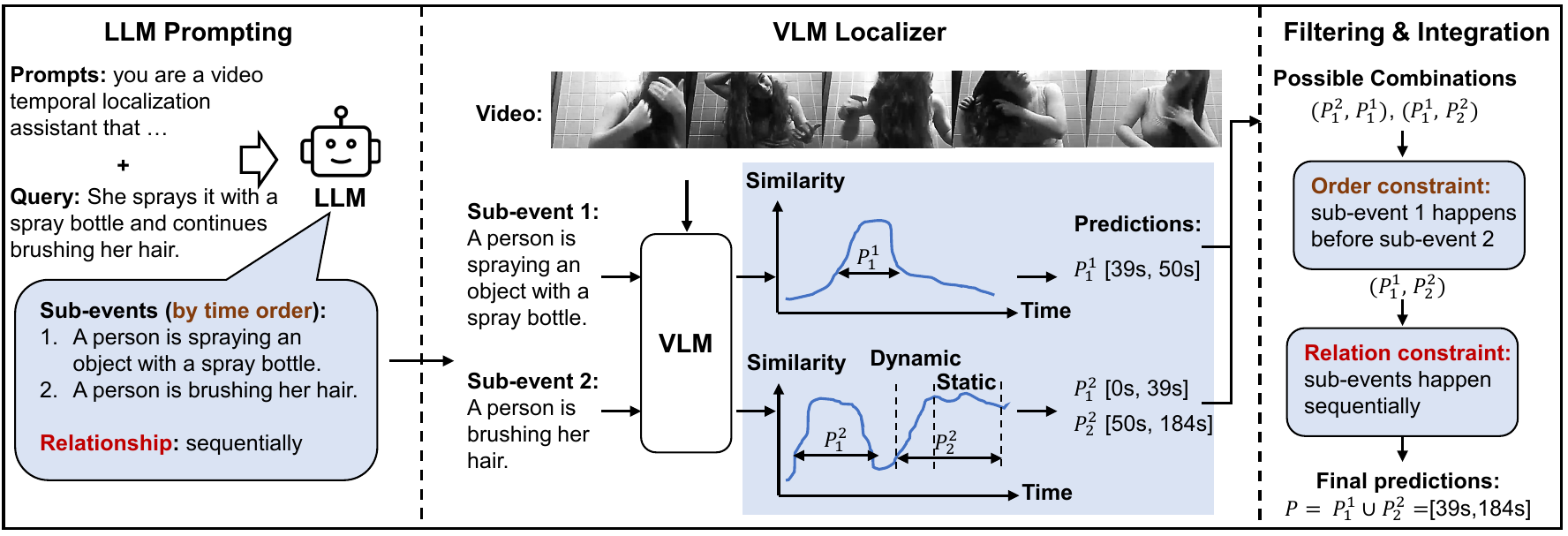}
    \caption{The pipeline of the proposed method. Firstly, the LLM prompting leverages the large language model (LLM) to analyze sub-events contained in the query and reason the order and the temporal relationship of these sub-events. Then, the VLM localizer uses the vision language models to localize the sub-event in the video. The VLM localizer calculates the similarity between the video frames and the sub-event descriptions, enumerates event proposals in the video, and explicitly considers both dynamic transition and static status post-transition when measuring the similarity between the proposal and the text query, thus selecting proposals as the localization results.   
    Finally, we filter and integrate the results of the VLM localizer based on the order and relationship of sub-events inferred by LLM to make the final prediction.}
    \label{fig:pipeline}
\end{figure}

\section{Method}

The overview of our model design is illustrated in Figure~\ref{fig:pipeline}. Our method consists of three parts: Firstly, since the query may describe multiple events that happened sequentially or simultaneously in the video, we propose the LLM prompting to leverage LLM for analyzing sub-events contained in the query and reason the order and the temporal relationship of these sub-events. Then, we propose the VLM localizer that uses the VLM to localize the sub-event in the video. The VLM localizer first calculates the similarity between the video frames and the sub-event descriptions. 
To solve the problem that VLMs are not sensitive enough to the dynamic transitions in the video, we propose explicitly considering both dynamic transitions and static states following these transitions when evaluating the similarity between the proposal and the text query. A good proposal should commence with a dynamic segment showing a significant rise in video-text similarity, followed by a static post-transition segment characterized by a high average video-text similarity within the static segment, while maintaining a low average similarity outside of it. To assess each proposal, we enumerate the plausible dynamic and static segments of every event proposal and calculate a matching score by aggregating the dynamic and static scores. The dynamic score measures the rate of similarity change within its dynamic segment, and the static score evaluates the comparative similarity within and outside its static post-status segment.
The VLM localizer returns the top-k proposals with the highest sum of dynamic and static scores as the localization results. Finally, we filter and integrate the results of the VLM localizer based on the order and relationship of sub-events inferred by LLM to make the final prediction.

\subsection{LLM Prompting}

The large language model (LLM) demonstrates powerful capabilities in instruction following, context understanding, and reasoning. Therefore, we propose to prompt LLM to analyze query texts, identifying multiple potential events therein, and analyzing the order and relationship of these events. Specifically, we request the large language model to provide the following information:
\begin{itemize}
    \item Reasoning: Analyze the user's query and infer the sub-events it may contain. 
    \item Order of sub-events: Provide each sub-events in chronological order.
    \item Relationships between sub-events: Consider three types of relationships: Single event, simultaneously (i.e. the sub-events occur simultaneously), and sequentially (i.e. the sub-events occur sequentially).
    \item Textual descriptions: Generate text descriptions for each sub-event. 
\end{itemize}
The complete prompt will be provided in the supplementary materials. 

\subsection{Grounding with Vison Language Model}
Inspired by the powerful multimodal alignment capabilities of VLM, we propose to use VLM as a localizer to locate each sub-event in the video. 

Specifically, we choose BLIP-2 Q-Former~\cite{li2023blip} following \cite{zhang2023videollama,li2023videochat} as the VLM localizer. For a sub-event description $c$ and each video frame $v_1,...,v_N$, we use VLM to extract their text features $f^c \in \mathbb{R}^{D}$ and vision features $F^v=[f^v_1,...,f^v_N]\in \mathbb{R}^{N\times D}$, where $N$ is the number of video frames and $D$ is the feature dimension. As the text and vision space are well aligned, we can directly use the cosine similarity of the text and vision features to measure the relevance of the sub-event description and the video frame:
\begin{equation}
    S = \frac{f^{c}F^{v\intercal}}{\|f^{c}\|\|F^{v}\|}\in \mathbb{R}^{N}
    \label{eq:sim}
\end{equation}

We find that directly enumerating proposals within the video and selecting the proposal with the highest average similarity often leads the model to only predict static status in the event while ignoring the dynamic transition at the beginning of an event. In 56.1\% of the testing data in the Charades-STA dataset, the naive baseline predicts a start timestamp that is after the ground truth start timestamp. 
For example, for the query "a person sits down", the model tends to predict segments where the person is already seated on the chair rather than the process of the person gradually from standing up to sitting down.
A good event should commence with a dynamic segment showing a significant rise in video-text similarity, followed by a static post-transition segment characterized by a high average video-text similarity within the static segment, while maintaining a low average similarity outside of it. To assess each proposal, we enumerate the plausible dynamic and static segments of every event proposal and calculate a matching score by aggregating the dynamic and static scores. The dynamic score measures the rate of similarity change within its dynamic segment, and the static score evaluates the comparative similarity within and outside its static post-status segment:

\subsubsection{Dynamic Scoring}

Considering the continuity of the video, in the dynamic part of the target segment, the video content transitions from another event to the target event described by the query, so the relevance between the video frames and the query should quickly increase. Our dynamic scoring aims to provide quantitative scores for segments where the relevance between the video and the query increases rapidly. Firstly, to eliminate the influence of video jitter, we apply a Gaussian filter to the similarity $S$: $\hat{S}=G(S)$,
where $G(\cdot)$ is the Gaussian filter. We expect that the dynamic section contains as many parts as possible where the video-text correlation significantly increases. Therefore, we calculate the difference in similarity $\hat{S}$: $D_{i} = \hat{S}_{i} -\hat{S}_{i-1}$.
Given a proposal starts from the $i$-th frame and ends with the $k$-th frame, we require that all the differential values within the proposal exceed a certain threshold $\delta$. If this condition is met, we sum up the differential values in the proposal to obtain the dynamic score for that proposal:
\begin{equation}
S^{dynamic}_{i,k}=\left\{
\begin{aligned}
&\sum_{l=i}^k D_l, &{D_l > \delta ,\forall l \in [i,k]}\\
&0, &{otherwise} \\
\end{aligned}
\right.
\end{equation}
\subsubsection{Static Scoring}

Inspired by SPL~\cite{zheng2023SPL}, in the static part, we require that the most relevant event for a given query should satisfy the requirement that videos within the event have high relevance to the query and videos outside the event have low relevance to the query. Therefore, given a proposal starts from the $k$-th frame and ends with the $j$-th frame, we calculate the average similarity within the proposal and the average similarity outside the proposal, and use the difference between them as the static scores:
\begin{equation}
    S^{static}_{k,j} = \frac{1}{j-k}\sum_{l\in [k,j]}S_{l}-\frac{1}{N - (j-k)}\sum_{l\notin [k,j]}S_{l}
\end{equation}
where $N$ is the number of frames.

To localize the target video clip using dynamic and static scoring, we enumerate event proposals in the video. For each proposal $(i, j)$, due to the varying lengths of transition segments in different events, we enumerate all feasible timestamps $k$ where the transition just finished and divide the proposal into two parts: the dynamic part $(i, k)$ and static part $(k, j)$. 
We then calculate the sum of the dynamic score and static score, and select the maximum value as the score for this proposal:
\begin{equation}
    S^{final}_{i,j} = \max_{k=i}^j (S^{dynamic}_{i,k}+S^{static}_{k,j})
\end{equation}
Finally, we select the top-k proposal with the highest final score $S^{final}$ as the localization results of the VLM localizer.

\subsection{Prediction Filtering and Integration}

The VLM locator returns the top-k predictions for each sub-event description. To obtain the final prediction, we propose to filter and integrate these predictions from the VLP localizer based on the order of occurrence of sub-events and their relationships derived from LLM. 
Firstly, for each sub-event, we enumerate one of its top-k predictions, and there are total $k^m$ combinations, where $m$ is the number of sub-events.
For a combination $P_1,P_2,...,P_m$, we filter these combinations by the order constraint: if LLM determines that $P_i$ from $s^i$ to $e^i$ should occur before $P_j$ from $s^j$ to $e^j$, but the start time of $P_i$ is later than the end time of $P_j$ (i.e. $s^i>e^j$), then this combination will be discarded. After filtering, we can calculate the sum of scores $S^{final}$ returned by the VLM localizer for each combination, selecting the combination with the highest score, and merging these proposals based on the relation constraint: If LLM determines that these sub-events should occur simultaneously, we take the intersection of these proposals as the final prediction; otherwise, we take the union of these proposals as the final prediction:
\begin{equation}
    P^{final}=\left\{
\begin{aligned}
&P_1\cap P_2\cap ...\cap P_m, &{\text{relation is `simultaneously'}}\\
&P_1\cup P_2\cup ...\cup P_m, &{\text{relation is `sequentially'}} \\
\end{aligned}
\right.
\end{equation}

%% file: secs/4_experiments.tex
\section{Experiments}

To comprehensively validate the effectiveness of our method, we conduct experiments on the Charades-STA~\cite{gao2017tall} and ActivityNet Captions~\cite{2017Dense} datasets. We compare our method with existing methods under the IID, OOD, and cross-dataset settings respectively to demonstrate the generalization capability of our method. We also conduct ablation studies to evaluate the effectiveness of each module.

\subsection{Experimental Setup}
\textbf{Dataset:} We conduct experiment on two benchmark ActivityNet Captions~\cite{2017Dense} and  Charades-STA~\cite{gao2017tall}. \textit{ActivityNet Captions} contains 20K videos, which is originally collected for video captioning. There are 37,417/17,505/17,031 video-query pairs in the train /val\_1/val\_2 split. We follow previous works~\cite{luo2024zero,MMN} and report the performance on the val\_2 split. 
\textit{Charades-STA} is built based on the Charades dataset. There are 12,408/3,720 video-query pairs in the train/test split. We report the performance on the test split.

\noindent \textbf{Evaluation Metrics:}
We follow the evaluation metrics ‘R@m’ and ‘mIoU’ in the previous work~\cite{luo2024zero,MMN}, where m is the predefined temporal Intersection over Union (IoU) threshold. The metric ‘R@m’ represents the percentage of predictions that have the IoU larger than m. The metric ‘mIoU’ represents the average IoU of all the predictions.

\noindent \textbf{Implementation Details:}
We follow \cite{zhang2023videollama,li2023videochat} to use the BLIP-2 Q-former~\cite{li2023blip} as the vision-language model. For the large language model, we use the GPT-4 Turbo API. 
For the VLM localizer, we downsample the input video to 3 FPS and use VLM to calculate the similarity between video frames and text. The localizer returns $k=3$ predictions for each sub-event. For the hyperparameter, we set $\delta$ to $5\times10^{-4}$ across all the datasets.

\subsection{Comparison with the SOTAs}

\textbf{Comparison under the IID setting.} 
In Table~\ref{tab:iid}, we compare the performance of our method with existing methods under the IID setting. We use the official splits of Charades-STA and ActivityNet Captions. It can be observed that our method achieves the best performance in the zero-shot setting. For example, on the Charades-STA dataset, our method surpasses the second-ranked VTG-GPT~\cite{xu2024vtggpt} by 6.29\% on the R@0.5 metric. Our method also outperforms unsupervised training methods by 9.27\% on the R@0.5 metric on the Charades-STA dataset. The methods that utilize both LLM and VLM, such as VideoLLaMA~\cite{zhang2023videollama} which aligns visual features to the input token space of LLM, have a poor performance on video temporal grounding. Although VTimeLLM~\cite{huang2023vtimellm} and GroundingGPT~\cite{li2024groundgpt} further finetune LLM using data from ActivityNet-Captions or Charades-STA, the performance remains suboptimal. Our method combines the advantages of LLM in text understanding and reasoning with the advantages of VLM in visual-text alignment, resulting in superior performance.

\textbf{Comparison under the OOD setting.} 
To study the generalization capability of our method, we conducted experiments under OOD settings. We consider three OOD settings: novel location, novel text, and cross-dataset.

For the novel location, we follow DCM~\cite{yang2021deconfounded} by inserting a segment of random-generated video at the beginning of test videos, as shown in Table~\ref{tab:ood}. In Table~\ref{tab:ood2}, we also test the performance on the Charades-CD~\cite{yuan2021closer} dataset, which alters the distribution of target location by resplitting the training and test data. As we can see, our method outperforms recent fully supervised methods in this setting without training, proving its superior generalization capability.

For the novel text, we follow VISA~\cite{li2022compositional} and test on the Charades-CG~\cite{li2022compositional} dataset as shown in Table~\ref{tab:ood2}. `Novel-composition' indicates the text contains novel combinations of training vocabulary, while `novel-word' indicates the text contains novel words. Our method achieves the best performance under the novel-word setting. Under the novel-composition setting, although not as competitive as methods specifically designed for this scenario, such as DeCo and VISA, our method still outperforms other fully supervised approaches.

For the cross-dataset setting, we follow Debias-TLL~\cite{bao2022learning} where the models are trained on ActivityNet Captions and tested on the Charades-STA, as shown in Table~\ref{tab:cross_dataset}. Notably, almost all fully supervised methods experience significant performance degradation when applied across datasets, while our method remains unaffected as it does not rely on training data distribution. These experiments demonstrate that our method has superior generalization capabilities, making it more suitable for practical application requirements.

\begin{table*}[t]
    \centering
    \scalebox{0.8}{
    \begin{tabular}{c|ccc|cccc|cccc}
    \toprule
        \multirow{2}{*}{Method} &\multirow{2}{*}{Setting} &\multirow{2}{*}{VLM}& \multirow{2}{*}{LLM} & \multicolumn{4}{c|}{Charades-STA} &\multicolumn{4}{c}{ActivityNet Captions} \\ 
         & && & R@0.3 & R@0.5 & R@0.7 & mIoU &R@0.3 & R@0.5 & R@0.7 & mIoU \\ 
    \midrule
        2D-TAN~\cite{zhang2020learning}& \multirow{4}{*}{fully}&\multirow{4}{*}{\XSolidBrush}&\multirow{4}{*}{\XSolidBrush}& - & 39.81 & 23.25 & -& 58.75 & 44.05 & 27.38 & - \\ 
        EMB~\cite{huang2022video}& & &  & \textbf{72.50} & 58.33 & 39.25 & \textbf{53.09}& \textbf{64.13} & 44.81 & 26.07 & \textbf{45.59}\\
        MGSL-Net~\cite{liu2022memory}&&& & - & 63.98 & 41.03 & -& - & 51.87 & 31.42 & -\\ 
        EaTR~\cite{jang2023EaTR}&&& & - & \textbf{68.47} & \textbf{44.92} & -& - & \textbf{58.18} & \textbf{37.64} & -\\
    \hline\hline     
        CRM~\cite{huang2021cross}&\multirow{4}{*}{weakly}&\multirow{4}{*}{\XSolidBrush}&\multirow{4}{*}{\XSolidBrush}& 53.66 & 34.76 & 16.37 &- & 55.26 & 32.19 &-&-  \\ 
        CNM~\cite{CNM_2022_AAAI}&&&&60.39 & 35.43 & 15.45& -&55.68 & 33.33 &-& -\\ 
        CPL~\cite{CPL_2022_CVPR}&&& & 66.40 &49.24& 22.39 & -& 55.73 &31.37&- & -\\
         Huang et al.~\cite{huang2023weakly} & & & & \textbf{69.16}& \textbf{52.18}& \textbf{23.94} &\textbf{45.20} & \textbf{58.07} &\textbf{36.91}& - & \textbf{41.02} \\
    \hline\hline    
        Gao et al.~\cite{gao2021learning}& \multirow{5}{*}{unsup.~\tablefootnote{Some of them claim to be under the zero-shot setting. However, they still require video data for training. We follow \cite{luo2024zero} to classify them as unsupervised methods.}} & \multirow{5}{*}{\Checkmark} & \multirow{5}{*}{\XSolidBrush}  & 46.69 & 20.14 & 8.27 & -& 46.15 & 26.38 & 11.64 & - \\
        PSVL~\cite{nam2021zero}&& &  & 46.47 & 31.29& 14.17 &31.24 & 44.74 & 30.08 & 14.74 & 29.62\\
        PZVMR~\cite{wang2022prompt}&& &  & 46.83 & 33.21 & 18.51 & 32.62& 45.73 & 31.26 & \textbf{17.84} & 30.35 \\
        Kim et al.~\cite{kim2023language}&& &  & 52.95 & 37.24 & 19.33 & 36.05& 47.61 & \textbf{32.59} & 15.42 & 31.85 \\
        SPL~\cite{zheng2023SPL} & & & & \textbf{60.73} & \textbf{40.70} & \textbf{19.62} & \textbf{40.47} & \textbf{50.24} & 27.24 & 15.03 & \textbf{35.44}\\
    \hline\hline
    GroundingGPT~\cite{li2024groundgpt} & \multirow{2}{*}{fully~\tablefootnote{They use the data in the ActivityNet Captions or Charades-STA to finetune LLMs.}}& \multirow{2}{*}{\Checkmark} &\multirow{2}{*}{\Checkmark} & - & 29.6 &11.9 & -&-&-&-&-\\
    VTimeLLM-13B~\cite{huang2023vtimellm} & &&&\textbf{55.3} &\textbf{34.3} &\textbf{14.7} &\textbf{34.6}& \textbf{44.8} &\textbf{29.5} &\textbf{14.2} &\textbf{31.4} \\

    \hline\hline    
    VideoChat-7B~\cite{li2023videochat} &\multirow{7}{*}{zero-shot}& \Checkmark & \Checkmark  &9.0& 3.3& 1.3 &6.5& 8.8& 3.7& 1.5 &7.2 \\
    VideoLLaMA-7B~\cite{zhang2023videollama}&& \Checkmark& \Checkmark&10.4 &3.8 &0.9 &7.1 &6.9 &2.1 &0.8 &6.5 \\
    VideoChatGPT-7B~\cite{maaz2023videochatgpt}&&  \Checkmark & \Checkmark &20.0 &7.7 &1.7 &13.7& 26.4& 13.6& 6.1& 18.9\\ 
    Luo et al.~\cite{luo2024zero}& & \Checkmark & \XSolidBrush & 56.77 & 42.93& 20.13& 37.92 &48.28& 27.90& 11.57& 32.37\\
    VTG-GPT~\cite{xu2024vtggpt}& &\Checkmark & \Checkmark& 59.48& 43.68& \textbf{25.94}& 39.81 &47.13& 28.25& 12.84 &30.49\\
    Ours w/o LLM &  & \Checkmark & \XSolidBrush &65.46 &48.01 &22.07 &43.37 & 48.84 &26.64 &13.10 &33.61 \\
    Ours &  & \Checkmark & \Checkmark &\textbf{67.04} &\textbf{49.97} & 24.32 & \textbf{44.51} & \textbf{49.34} &	\textbf{27.02}	& \textbf{13.39} &\textbf{34.10} \\
    \bottomrule
    \end{tabular}}
    \caption{Evaluation Results on the Charades-STA and  ActivityNet Captions Datasets.}
    \label{tab:iid}
\end{table*}

\begin{table}[t]
    \centering
    \scalebox{0.7}{
    \begin{tabular}{c|c|ccc|ccc|ccc|ccc}
    \toprule
         \multirow{3}{*}{Method} & \multirow{3}{*}{Setting} & \multicolumn{6}{c|}{Charades-STA} &\multicolumn{6}{c}{ActivityNet-Captions}  \\
         & & \multicolumn{3}{c|}{OOD-1} & \multicolumn{3}{c|}{OOD-2} & \multicolumn{3}{c|}{OOD-1} & \multicolumn{3}{c}{OOD-2} \\
         & & R@0.5 & R@0.7 & mIoU & R@0.5 & R@0.7 & mIoU & R@0.5 & R@0.7 & mIoU & R@0.5 & R@0.7 & mIoU \\
    \midrule
         LGI~\cite{LGI}  & \multirow{5}{*}{fully} & 42.1 & 18.6 & 41.2 & 35.8 & 13.5 & 37.1 & 16.3 & 6.2 & 22.2 & 11.0 & 3.9 & 17.3 \\
         2D-TAN~\cite{zhang2020learning}  & & 27.1 & 13.1 & 25.7 & 21.1 & 8.8 & 22.5 & 16.4 & 6.6 & 23.2 & 11.5 & 3.9 & 19.4 \\
         MMN~\cite{MMN} & & 31.6 & 13.4 & 33.4 & 27.0 & 9.3 & 30.3 & 20.3 & 7.1 & 26.2 & 14.1 & \textbf{5.2} & 20.6 \\
         VDI~\cite{luo2023vid} & & 25.9 & 11.9 & 26.7 & 20.8 & 8.7 & 22.0 & \textbf{20.9} & 7.1 & \textbf{27.6} & \textbf{14.3} & \textbf{5.2} & \textbf{23.7} \\
         DCM~\cite{yang2021deconfounded} & & \textbf{44.4} & \textbf{19.7} & \textbf{42.3} & \textbf{38.5} & \textbf{15.4} & \textbf{39.0} & 18.2 & \textbf{7.9} & 24.4 & 12.9 & 4.8 & 20.7 \\
    \hline
        CNM~\cite{CNM_2022_AAAI} & \multirow{2}{*}{weakly} & 9.9 & 1.7 & 21.6 & 6.1 & 0.5 & 16.6 & \textbf{6.1} & \textbf{0.4} & 21.0 & \textbf{2.5} & 0.1 & 16.8 \\
        CPL~\cite{CPL_2022_CVPR} & & \textbf{29.9} & \textbf{8.5} & \textbf{32.2} & \textbf{24.9} & \textbf{6.3} & \textbf{30.5} & 4.7 & \textbf{0.4} & \textbf{21.1} & 2.1 & \textbf{0.2} & \textbf{17.7} \\
    \hline
        PSVL~\cite{nam2021zero} & \multirow{2}{*}{unsup.} & \textbf{3.0} & 0.7 & 8.2 & \textbf{2.2} & 0.4 & 6.8 & - & - & - & - & - & - \\
        PZVMR~\cite{wang2022prompt} & & - & \textbf{8.6} & \textbf{25.1} & - & \textbf{6.5} & \textbf{28.5} & - & \textbf{4.4} & \textbf{28.3} & - & \textbf{2.6} & \textbf{19.1} \\
    \hline
        Luo et al.~\cite{luo2024zero} & \multirow{2}{*}{zero-shot} & 40.3 & 18.2 & 38.2 & 38.9 & 17.0 & 37.8 & 18.4 & 6.8 & 21.1 & \textbf{18.6} & 7.4 & 20.6 \\
        Ours & &\textbf{45.9} &\textbf{20.8}&\textbf{43.0}&\textbf{43.8}&\textbf{20.0}&\textbf{42.6}&\textbf{20.4}&\textbf{11.2}&\textbf{31.7}&18.5&\textbf{10.0}&\textbf{30.3} \\
    \bottomrule
    \end{tabular}}
    \caption{Results under OOD setting on the Charades and  ActivityNet Dataset.}
    \label{tab:ood}
\end{table}

\begin{table}[t]
    \centering
    \scalebox{0.7}{
    \begin{tabular}{c|c|ccc|ccc|ccc}
    \toprule
         \multirow{3}{*}{Method} & \multirow{3}{*}{Setting} & \multicolumn{3}{c|}{Charades-CD} &\multicolumn{6}{c}{Charades-CG}  \\
         & & \multicolumn{3}{c|}{test-ood} & \multicolumn{3}{c|}{novel-composition} & \multicolumn{3}{c}{novel-word} \\
         & & R@0.3 & R@0.5 & R@0.7 & R@0.5 & R@0.7 & mIoU & R@0.5 & R@0.7 & mIoU \\
    \midrule
        2D-TAN~\cite{zhang2020learning} &\multirow{5}{*}{fully} & 43.45 & 30.77 & 11.75 & 30.91 & 12.23 & 29.75 & 29.36 & 13.21 & 28.47 \\
        TSP-PRL~\cite{wu2020tree} & & 31.93 & 19.37 & 6.20 & 16.30 & 2.04 & 13.52 & 14.83 & 2.61 & 14.03 \\
        SCDM~\cite{yuan2019semantic} &  & \textbf{52.38} & \textbf{41.60} & \textbf{22.22} & 27.73 & 12.25 & 30.84 &- & -&- \\
        VISA~\cite{li2022compositional} &  & - & - & - & 45.41 & \textbf{22.71} & \textbf{42.03} & \textbf{42.35} & \textbf{20.88} & \textbf{40.18} \\
        DeCo~\cite{yang2023deco} & & - & - & - & \textbf{47.39} & 21.06 & 40.70 & - & -  & -  \\
    \midrule
        WSSL~\cite{duan2018weakly} & \multirow{2}{*}{weakly} & \textbf{35.86} & \textbf{23.67} & \textbf{8.27} & 3.61 & 1.21 & 8.26 & 2.79 & 0.73 & \textbf{7.92} \\
        CPL~\cite{CPL_2022_CVPR} &  & - & - & - & \textbf{39.11} & \textbf{15.60} & \textbf{35.53} &  \textbf{45.90}
&\textbf{22.88} &-   \\
    \midrule
        SPL~\cite{zheng2023SPL} & \multirow{1}{*}{unsup.} & 62.96 & 38.25 & 15.53 & - & -& -&- &-&-\\
    \midrule
        Luo et al.~\cite{luo2024zero} & \multirow{2}{*}{zero-shot} &- & -& -& 40.27 &16.27 & -&45.04& 21.44 & -\\
        Ours & &\textbf{65.07}&\textbf{49.24} &\textbf{23.05}&\textbf{43.84} &\textbf{18.68} & \textbf{40.19}&\textbf{56.26}&\textbf{28.49}&\textbf{46.90}\\
    \bottomrule
    \end{tabular}}
    \caption{Results under OOD setting on the Charades-CD and  Charades-CG Dataset.}
    \label{tab:ood2}
\end{table}

\begin{table}[t]
\begin{minipage}[t]{0.48\linewidth}
    \centering
    \scalebox{0.7}{
    \begin{tabular}{c|cccc}
    \toprule
         \multirow{2}{*}{Method} & R@1 & R@1 & R@5 & R@5 \\
         & R@0.5 & R@0.7 & R@0.5 & R@0.7 \\
    \midrule
         SCDM~\cite{yuan2019semantic} & 15.91 & 6.19 & 54.04 & 30.39 \\
         2D-TAN~\cite{zhang2020learning} & 15.81 & 6.30 & 59.06 & 31.53 \\
         Debias-TLL~\cite{bao2022learning} & 21.45 & 10.38 & 62.34 & 32.90 \\
         Ours & \textbf{49.97} & \textbf{24.32} &  \textbf{83.5} & \textbf{42.2}  \\
    \bottomrule
    \end{tabular}}
    \caption{Cross-dataset performance when training on ActivityNet captions and evaluate on Charades-STA.}
    \label{tab:cross_dataset}
\end{minipage}
\hspace{2mm}
\begin{minipage}[t]{0.48\linewidth}
    \centering
    \scalebox{0.7}{
    \begin{tabular}{cccc|ccc}
    \toprule
          &\makecell{LLM\\prompting} & \makecell{VLM\\localizer} & \makecell{Filtering \&\\ Integration} & R@0.5 & R@0.7 & mIoU \\
    \midrule
       1&\XSolidBrush &\XSolidBrush &\XSolidBrush  &42.32& 18.91& 31.61 \\
    \midrule
         2&\Checkmark &&&  43.17& 18.56 & 32.14\\
         3&\Checkmark &&\Checkmark & 44.12& 19.21 &33.07 \\
         4& &\Checkmark&& 48.01 & 22.07 &43.37 \\
         5&\Checkmark &\Checkmark&&  48.41 & 21.94 & 42.76\\
    \midrule
        6&\Checkmark &\Checkmark&\Checkmark& \textbf{49.97} & \textbf{24.32} & \textbf{44.51} \\
    \bottomrule
    \end{tabular}}
    \caption{Ablations on each component.}
    \label{tab:ab_comp}
\end{minipage}
\end{table}

\begin{table}
\begin{minipage}[t]{0.48\linewidth}
    \centering
    \scalebox{0.7}{
    \begin{tabular}{cc|ccc}
    \toprule
          \makecell{Dynamic\\Scoring} & \makecell{Static\\Scoring} & R@0.5 & R@0.7 & mIoU \\
    \midrule
          \XSolidBrush&\XSolidBrush& 42.32 &18.91 &31.61 \\
    \midrule
          \Checkmark&& 47.63& 20.13&41.68 \\
          &\Checkmark& 45.48 & 22.02 &41.81 \\
    \midrule
         \Checkmark&\Checkmark& \textbf{48.01} & \textbf{22.07} & \textbf{43.37}\\
    \bottomrule
    \end{tabular}}
    \caption{Ablations on VLM localizer.}
    \label{tab:ab_vlm}
\end{minipage}
\hspace{2mm}
\begin{minipage}[t]{0.48\textwidth}
    \centering
    \scalebox{0.7}{
    \begin{tabular}{cc|ccc}
    \toprule
          \makecell{Order\\ Constraint} & \makecell{Relation \\Constraint} & R@0.5 & R@0.7 & mIoU \\
    \midrule
          \XSolidBrush&\XSolidBrush& 42.32 & 18.91& 31.61 \\
    \midrule
          \Checkmark& & 43.01 & 19.03 & 31.73\\
          &\Checkmark& 43.97& 19.11 & 32.76\\
    \midrule
         \Checkmark&\Checkmark& \textbf{44.12}& \textbf{19.21} &\textbf{33.07} \\
    \bottomrule
    \end{tabular}}
    \caption{Ablations on LLM prompting.}
    \label{tab:ab_llm}
\end{minipage}
\end{table}

\begin{table}
\begin{minipage}[t]{0.48\textwidth}
    \centering
    \scalebox{0.7}{
    \begin{tabular}{cc|ccc}
    \toprule
          VLMs& Type & R@0.5 & R@0.7 & mIoU \\
    \midrule
    CLIP~\cite{radford2021learning} &  \multirow{2}{*}{Image}& 42.68& 18.92 & 38.89 \\
         BLIP-2~\cite{li2023blip}&&\textbf{48.01} & \textbf{22.07}& \textbf{43.37}\\
    \midrule
         InterVideo~\cite{wang2022internvideo}&\multirow{2}{*}{Video}& 44.60 & 20.51 &40.72 \\
         ViCLIP~\cite{wang2023internvid}& &44.01 & 20.48 &40.25 \\
    \bottomrule
    \end{tabular}}
    \caption{Ablations on the VLMs.}
    \label{tab:ab_vlm_model}
\end{minipage}
\hspace{2mm}
\begin{minipage}[t]{0.48\textwidth}
    \centering
    \scalebox{0.7}{
    \begin{tabular}{c|ccc}
    \toprule
          LLMs & R@0.5 & R@0.7 & mIoU \\
    \midrule
         None &48.01 & 22.07& 43.37\\
         Gemini-1.0-Pro~\cite{team2023gemini} &48.97& 22.76 &44.12 \\
         GPT-3.5 Turbo  &49.23& 23.11 &\textbf{44.69} \\
         GPT-4 Turbo  & \textbf{49.97} & \textbf{24.32} & \textbf{44.51} \\
    \bottomrule
    \end{tabular}}
    \caption{Ablations on the LLMs.}
    \label{tab:ab_llm_model}
\end{minipage}

\end{table}

\subsection{Ablation Studies}

To validate the effectiveness of each module, we conduct ablation studies on the Charades-STA dataset.

\textbf{Effectiviness of each component.}
Table~\ref{tab:ab_comp} shows the ablation studies on the effectiveness of the proposed modules. 
When disenabling the VLM localizer, we use the naive baseline. When disenabling Filtering\&Integration, we simply take the top-1 predictions of the localizer for each sub-event and consider their union box as the final prediction. 
(1) From the 2nd row of the table, it can be observed that using LLM alone without Filtering\&Integration may hurt some metrics. This is because the descriptions of sub-events usually only capture a part of the semantics of the original query, and the localizing results are inaccurate without Filtering\&Integration. (2) From the 3rd row of the table, it can be seen that when both LLM prompting and Filtering\&Integration are used, the model outperforms the naive baseline by 1.46\% in mIoU. (3) From the 4th row of the table, our proposed VLM localizer shows a significant performance improvement, with a 5.69\% increase in R@0.5 compared to the naive baseline. (4) In the 6th row, when all three modules are enabled, performance is further improved, demonstrating the effectiveness of our method.

\textbf{Effectiviness of VLM.}
Table~\ref{tab:ab_vlm} shows the effectiveness of our two scoring functions in the VLM Localizer. (1) From the second row of the table, it can be observed that the dynamic scoring significantly improves performance, with an increase of 10.07\% in mIoU. Since most VLMs are trained on image-text or trimmed video clip-text data, they are not sensitive enough to the dynamic transitions between different events in the same video. Dynamic scoring models the dynamic transitions implicitly, thus demonstrating better performance. (2) From the third row of the table, when using static scoring, the mIoU of the naive baseline improves by 10.2\%. Static scoring compared with the naive baseline not only requires high visual-textual relevance within the event but also requires low visual-textual relevance outside the event, thereby avoiding the model's focus solely on the most discriminative video segments. Combining both approaches further enhances model performance, demonstrating the effectiveness of our VLM localizer.

In Table~\ref{tab:ab_vlm_model}, we report the performance of different VLMs, including the image-text pre-trained model (CLIP~\cite{radford2021learning} and BLIP-2~\cite{li2023blip}) and video-text pre-trained model (InterVideo~\cite{wang2022internvideo}, ViCLIP~\cite{wang2023internvid}). It can be observed that BLIP-2 exhibits the best performance, even surpassing models trained on video-text data. We attribute this to the fact that the pretraining data for image-text is much larger than that for video-text (e.g. LAION400M~\cite{schuhmann2021laion} v.s. WebVid10M~\cite{bain2021frozen}), thus BLIP-2 demonstrates better generalization capability. Additionally, our designed dynamic scoring helps BLIP-2 better understand the dynamic transition in the videos. Notably, in Table~\ref{tab:iid}, VideoLLaMA~\cite{zhang2023videollama} and VideoChat~\cite{li2023videochat} utilize the frozen BLIP-2 Q-Former. VTG-GPT employs MiniGPT~\cite{zhu2023minigpt}, which is also based on the frozen BLIP-2. Therefore, our comparison with them is fair. Additionally, Luo et al.~\cite{luo2024zero} use InterVideo as the VLM, and as shown in Table~\ref{tab:ab_vlm_model}, our performance using InterVideo still surpasses them. 

\textbf{Effectiviness of LLM.} 
(1) We utilize the order and relationships of sub-events provided by LLM to filter and integrate the predictions of the VLM localizer. Table~\ref{tab:ab_llm} verifies the effectiveness of Filtering\&Integration. It can be observed that both order and relation constraints improve the performance, with the most significant improvement when both are used simultaneously. (2) In Table~\ref{tab:ab_llm_model}, we also report the performance of different LLMs, including Gemini-1.0-Pro, GPT-3.5 Turbo, and GPT-4 Turbo. The results indicate that more powerful LLM (e.g. GPT-4) can lead to better performance.

\begin{figure}
    \centering
    \includegraphics[width=\linewidth]{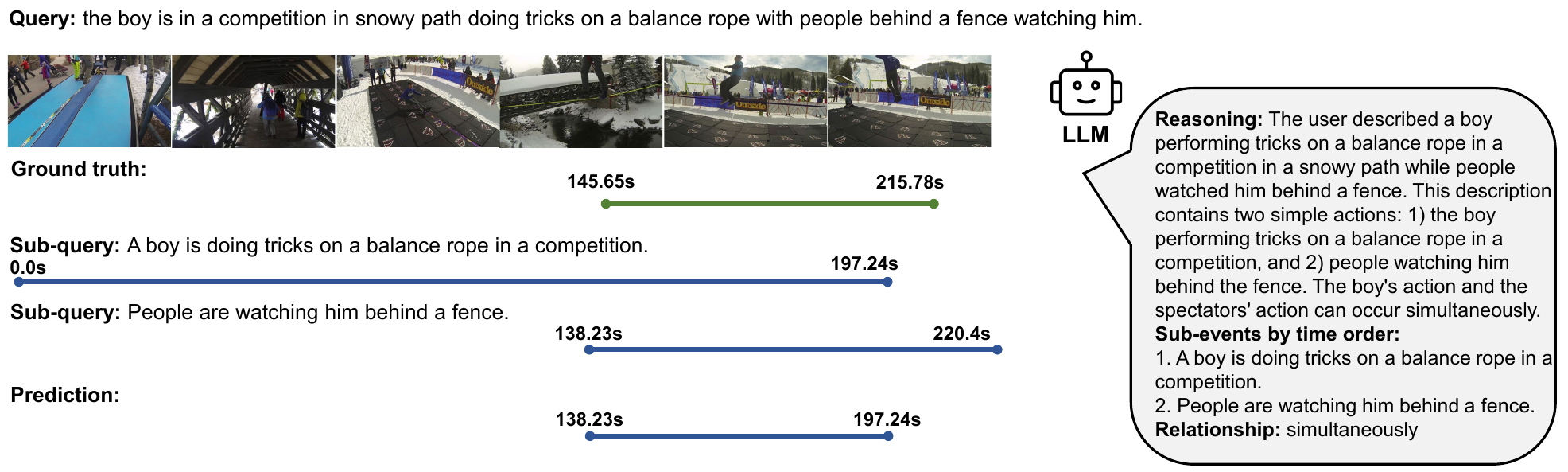}
    \caption{Qualitative results on the ActivityNet Captions dataset.}
    \label{fig:vis}
\end{figure}

\subsection{Qualitative Results}

Figure~\ref{fig:vis} presents a visualization. It can be seen that the LLM successfully splits the query into two sub-events and analyzes that both of these sub-events should occur simultaneously in the target query. Our VLM localizer successfully localizes these two sub-events, and their intersection forms the final prediction. More qualitative results can be found in the supplementary materials. 

%% file: secs/5_conclusion.tex
\section{Conclusion}

In this work, we study the problem of training-free video temporal grounding. We leverage the ability of LLM and VLM, requiring no specific video temporal localization dataset. We propose leveraging LLM to analyze multiple sub-events contained in the query and analyze the temporal order and relationships between these events. Then, we explicitly model the dynamic transition and static status in the video and use the VLM to localize the sub-events and leverage the order and relationships provided by LLMs to integrate the predictions.  Our method achieves the best performance on zero-shot video temporal grounding on Charades-STA and ActivityNet Captions datasets without any training and demonstrates better generalization in cross-dataset and OOD settings.

\textbf{Limitations.} LLM are not always reliable in reasoning the order and relationships between sub-events, which can negatively impact the performance. How to validate the reliability of outputs from LLM can be studied in the future.

\section*{Acknowledgement} 
This work was supported by grants from the National Natural Science Foundation of China (62372014, 61925201, 62132001, U22B2048).